\documentclass{article}

\usepackage{microtype}
\usepackage{graphicx}
\usepackage{booktabs} 

\usepackage{amssymb}
\usepackage{amsthm,amsmath}

\usepackage{hyperref}

\DeclareMathOperator\erf{erf}
\usepackage[accepted]{icml2018}

\icmltitlerunning{Accuracy-Reliability Cost Function}

\begin{document}

\twocolumn[
\icmltitle{Accuracy-Reliability Cost Function for Empirical Variance Estimation}




\begin{icmlauthorlist}
\icmlauthor{Enrico Camporeale}{to}

\icmlaffiliation{to}{Centrum Wiskunde \& Informatica, Multiscale Dynamics, Amsterdam, The Netherlands}

\icmlcorrespondingauthor{Enrico Camporeale}{e.camporeale@cwi.nl}
\end{icmlauthorlist}

\icmlkeywords{Machine Learning, ICML}

\vskip 0.3in
]



\printAffiliationsAndNotice{}  

\begin{abstract}
In this paper we focus on the problem of assigning uncertainties to single-point predictions.
We introduce a cost function that encodes the trade-off between accuracy and reliability in probabilistic forecast. We derive analytic formula for the case of forecasts of continuous scalar variables expressed in terms of Gaussian distributions. The Accuracy-Reliability cost function can be used to empirically estimate the variance in heteroskedastic regression problems (input-dependent noise), by solving a two-objective optimization problem. The simple philosophy behind this strategy is that predictions based on the estimated variances should be both accurate and reliable (i.e. statistical consistent with observations). We show several examples with synthetic data, where the underlying hidden noise function can be accurately recovered, both in one and multi-dimensional problems. The practical implementation of the method has been done using a Neural Network and, in the one-dimensional case, with a simple polynomial fit.
\end{abstract}

\section{Introduction}
There is a growing consensus, across many fields and applications, that forecasts should have a probabilistic nature \cite{gneiting14}. This is particularly true in decision-making scenarios where cost-loss analysis are designed to take into account the uncertainties associated to a given forecast \cite{murphy77, owens14}. Unfortunately, it is often the case that well established predictive models are completeley deterministic and thus provide single-point estimates only. In several domains of applied physics, where models rely on computer simulations, a typical strategy to assign confidence intervals to deterministic predictions is to perform ensemble forecasting \cite{gneiting05, leutbecher08}. However, this is expensive and it often requires a trade-off between complexity and accuracy of the model, especially when there is a need for real-time predictions.\\
In this paper we focus on the problem of assigning uncertainties to single-point predictions. We restrict our attention on predictive models that output a scalar continuous variable, and whose uncertainties are in general input-dependent. For the sake of simplicity, and for its widespread use, we assume that the probabilistic forecast that we want to generate, based on single-point outputs, is in the form of a Gaussian distribution. Hence, the problem can be recast in terms of the estimation of the empirical variance associated to a normal distribution centered around forecasted values provided by a model. Even though we will embrace a viewpoint stemming on the practical application of producing usable forecasts, this problem is very similar and technically equivalent to a particular case of heteroskedastic regression. Indeed, in the regression literature it is common to assume that noisy targets are generated as $y = f(\mathbf{x}) + \varepsilon$. Here we restrict to the special case of zero-mean Gaussian noise $\varepsilon \sim \mathcal{N}(0,\sigma^2(\mathbf{x}))$, where the heteroskedasticity is evident in the $\mathbf{x}$ dependence of the variance. Moreover, in order to bridge between the two viewpoints, we assume that the mean function $f(\mathbf{x})$ is provided by a black-box model, hence the focus is completely shifted on the estimation of the variance.
In the Machine Learning community, elegant and practical ways of deriving uncertainties based on Bayesian methods are well established, either based on Bayesian neural networks \cite{mackay92, neal12, hernandez15}, deep learning \cite{gal16}, or Gaussian Processes (GPs) \cite{rasmussen06}. The latter are successfully used in a variety of applications, their appeal being the non-parametric nature and their straightforward derivation based on linear algebra.
The standard derivation of GPs assumes input-independent (homoskedastic) target noise. \citet{goldberg98} have introduced an heteroskedastic GP model where the noise variance is derived with a second Gaussian Process, in addition to the GP governing the noise-free output values, and the posterior distribution of the noise is sampled via Markov chain Monte Carlo. Building on that work, several authors have proposed to replace the full posterior calculated through MCMC with maximum a posteriori (MAP) estimations \cite{kersting07, quadrianto09}, variational approximation \cite{lazaro11}, expectation propagation \cite{munoz2011}. See also \citet{le05,boukouvalas09,binois2016,antunes17}.\\
In this paper we embrace a complementary approach, focusing on parametric models for the empirical estimation of the variance. Our method is very general and decoupled from any particular choice for the model that predicts the output targets. The philosophy is to introduce a cost function, which encodes a trade-off between the accuracy and the reliability of a probabilistic forecast. Assessing the goodness of a forecast through standard metrics, such as the Continuous Rank Probability Score, is a common practice in many applications, like weather predictions \cite{matheson76, brocker07}. Also, the notion that a probabilistic forecast should be well calibrated, or statistically consistent with observations, has been discussed at length in the atmospheric science literature \cite{murphy92, toth03}. However, the basic idea that these two metrics (accuracy and reliability) can be exploited to estimate the empirical variance from a sample of observations, and possibly to reconstruct the underlying noise as a function of the inputs has not been explored yet.
In short, the empirical variances on a set of training points are found as the result of a multidimensional optimization problem that minimizes our cost function. In order to benchmark the method we show two simple implementations for deriving the estimated variance as a function of the inputs, and allowing to calculate the unknown variance on new test points: the first is based on Neural Network and the second on low-dimensional polynomial basis. 

\begin{figure}[ht]
\vskip 0.1in
\begin{center}
\centerline{\includegraphics[width=\columnwidth]{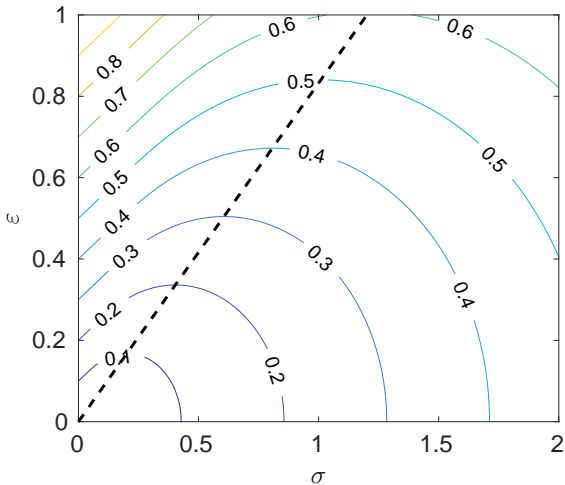}}
\caption{Lines of constant CRPS in $(\sigma,\varepsilon)$. The value of CRPS is indicated on the isolines. The black dashed line shows the location of $\sigma_{\text{min}}$ (i.e. the smallest CRPS for a given $\varepsilon$).}
\label{fig:CRPS}
\end{center}
\vskip -0.1in
\end{figure}

\section{Continuous Rank Probability Score}
The Continuous Rank Probability Score (CRPS) is a generalization of the well-known Brier score \cite{wilks11}, used to assess the probabilistic forecast of continuous scalar variables, when 
the forecast is given in terms of a probability density function, or its cumulative distribution. CRPS is defined as 
\begin{equation}
  \text{CRPS} = \int_{-\infty}^\infty \left[P(y) - H(y-y^o) \right]^2 dy
\end{equation}
where $P(y)$ is the cumulative distribution (cdf) of the forecast, $H(y)$ is the Heaviside function, and $y^o$ is the true (observed) value of the forecasted variable.
CRPS is a negatively oriented score: it is unbounded and equal to zero for a perfect forecast with no uncertainty (deterministic). 
In this paper we restrict our attention to the case of probabilistic forecast in the form of Gaussian distributions. Hence, a forecast is simply given by the mean value $\mu$
and the variance $\sigma^2$ of a Normal distribution. In this case $P(y) = \frac{1}{2}\left[\erf\left(\frac{y-\mu}{\sqrt{2}\sigma}\right) + 1\right]$
and the CRPS can be calculated analytically \cite{gneiting05} as

\begin{multline}\label{CRPS}
\text{CRPS}(\mu,\sigma,y^o) = \sigma\left[\frac{y^o-\mu}{\sigma}\erf\left(\frac{y^o-\mu}{\sqrt{2}\sigma}\right) + \right.  \\ 
 \left.\sqrt{\frac{2}{\pi}}\exp\left(-\frac{(y^o-\mu)^2}{2\sigma^2} \right) -\frac{1}{\sqrt{\pi}}\right]
\end{multline}

Several interesting properties of the CRPS have been studied in the literature. Notably, its decomposition into reliability and uncertainty has been shown in \citet{hersbach00}. The CRPS has the same unit as the variable of interest, and it collapses to the Absolute Error $|y^o-\mu|$ for $\sigma\rightarrow 0$, that is when the forecast becomes deterministic. CRPS is defined for a single instance of forecast and observation, hence it is usually averaged over an ensemble of predictions, to obtain the score relative to a given model: $\overline{\text{CRPS}} = \sum_k \text{CRPS}(\mu_k,\sigma_k,y^o_k)$. 
If we approach the problem of variance estimation in the framework of assigning an empirical variance to predictions originally made as single-point estimates, than it makes sense to minimize the CRPS, as a function of $\sigma$ only, for a fixed value of the error $\varepsilon=y^0-\mu$.
By differentiating Eq.(\ref{CRPS}) with respect to $\sigma$, one obtains
\begin{equation}
 \frac{d \text{CRPS}}{d\sigma} = \sqrt{\frac{2}{\pi}}\exp\left(-\frac{\varepsilon^2}{2\sigma^2} \right) -\frac{1}{\sqrt{\pi}}
\end{equation}
and the minimizer is found to be
\begin{equation}\label{sigma_min}
 \sigma_{\text{min}}^{\text{CRPS}} = \frac{\varepsilon}{\sqrt{\log 2}}.
\end{equation}
The CRPS penalizes under- and over-confident predictions in a non-trivial way. Indeed, for any value of the error $\varepsilon$, there are always two values of $\sigma$ (one smaller and one larger than $\sigma_{\text{min}}$) that yield the same CRPS. We show in Figure \ref{fig:CRPS} the isolines of CRPS in $(\sigma,\varepsilon)$ space. The black dashed line indicates $\sigma_{\text{min}}$. From this Figure it is clear how a smaller error $\varepsilon$ (for constant $\sigma$) always results in a smaller score, but the same score can be achieved by changing both the error $\varepsilon$ and the standard deviation $\sigma$.

\begin{figure}[!h]
\vskip 0.1in
\begin{center}
\centerline{\includegraphics[width=\columnwidth]{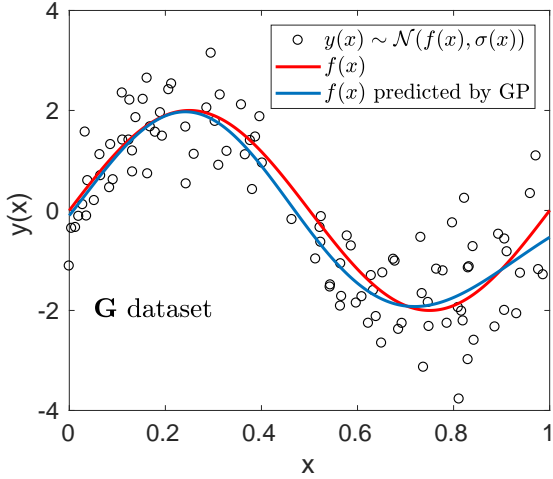}}
\caption{100 points sampled from the {\bf G} dataset (circles). The blue line shows the true mean function $f(x)$, while the red one is the one predicted by the GP model.}
\label{G_regression}
\end{center}
\vskip -0.1in
\end{figure}

\section{Reliability Score for Gaussian forecast}
Reliability is the property of a probabilistic model that measures its statistical consistence with observations. In particular, for forecasts of discrete events, the reliability measures if an event predicted with probability $p$ occurs, on average, with frequency $p$. 
This concept can be extended to forecasts of a continuous scalar quantity by examining the so-called rank histogram \cite{anderson96, hamill97, hamill01}.
\citet{hersbach00} focused on the case of a probabilistic forecast defined as the empirical distribution produced by an ensemble of deterministic predictions.
Contrary to the CRPS, that is defined for a single pair of forecast-observation, it is clear that the reliability can only be defined for a large enough ensemble of such pairs. 
Here, we introduce the reliability score for normally distributed forecasts. In this case, we expect the (scaled) relative errors $\eta=\varepsilon/(\sqrt{2}\sigma)$ calculated over a sample of $N$ predictions-observations to have a standard normal distribution with cdf $\Phi(\eta)=\frac{1}{2}(\erf(\eta)+1)$. Hence we define the Reliability Score (RS) as:
\begin{equation}\label{RS_1}
 \text{RS} = \int_{-\infty}^\infty \left[\Phi(\eta) - C(\eta)\right]^2 d\eta
\end{equation}
where $C(\eta)$ is the empirical cumulative distribution of the relative errors $\eta$, that is 
\begin{equation}
 C(y) = \frac{1}{N}\sum_{i=1}^N H(y-\eta_i)
\end{equation}
with $\eta_i = (y^o_i-\mu_i)/(\sqrt{2}\sigma_i)$. Obviously, RS measures the divergence of the empirical distribution of relative errors $\eta$ from a standard normal distribution. 
From now on we will use the convention that the set $\{\eta_1,\eta_2,\ldots \eta_N\}$ is sorted ($\eta_i\leq\eta_{i+1}$). Obviously this does not imply that $\mu_i$ or $\sigma_i$ are sorted as well.
Interestingly, the integral in Eq. (\ref{RS_1}) can be calculated analytically, via expansion into a telescopic series, yielding:
\begin{multline}\label{RS}
  \text{RS} = \sum_{i=1}^N \left[\frac{\eta_i}{N}\left(\erf(\eta_i)+1\right) - \frac{\eta_i}{N^2}(2i-1) \right.\\
  +\left. \frac{\exp(-\eta_i^2)}{\sqrt{\pi}N}\right] -\frac{1}{2}\sqrt{\frac{2}{\pi}}
\end{multline}
\normalsize
Differentiating the $i$-th term of the above summation, RS$_i$, with respect to $\sigma_i$ (for fixed $\varepsilon_i$), one obtains
\begin{equation}
 \frac{d\text{RS}_i}{d\sigma_i} =   \frac{\eta_i}{N\sigma_i}\left(\frac{2i-1}{N}-\erf(\eta_i)-1 \right)
\end{equation}
Hence, $\text{RS}_i$ is minimized at the value $\sigma_{\text{min}}^{\text{RS}}$ that satisfies
\begin{equation}\label{optimal_eta}
 \erf(\eta_i)=\erf\left(\frac{\varepsilon_i}{\sqrt{2}\sigma_{\text{min}}^{\text{RS}}}\right) = \frac{2i-1}{N}-1
\end{equation}
This could have been trivially derived by realizing that the distribution of $\eta_i$ that minimizes RS is the one such that the values of the empirical cumulative distribution $C(\eta)$ are uniform in the interval $[0,1]$.

\begin{figure}[!ht]
\begin{center}
\centerline{\includegraphics[width=\columnwidth]{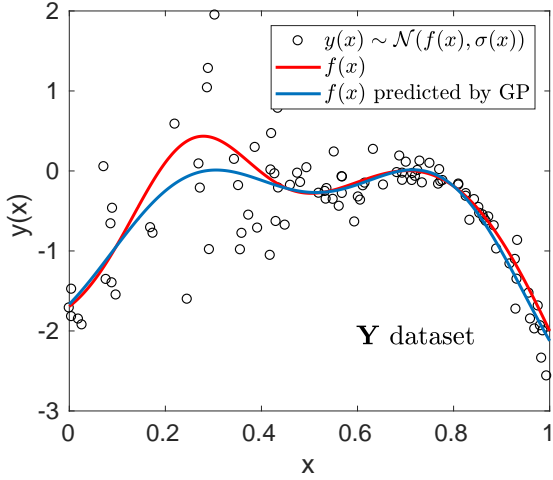}}
\caption{Same as in Figure \ref{G_regression}, for the {\bf Y} dataset.}
\label{Y_regression}
\end{center}
\vskip -0.1in
\end{figure}

\section{The Accuracy-Reliability cost function}
The Accuracy-Reliability cost function introduced here follows from the simple principle that the empirical variances $\sigma_i^2$ estimated from an ensemble of errors $\varepsilon_i$ should result in a model that is both accurate (e.g., with respect to the CRPS score), and reliable (e.g., with respect to the RS score). Clearly, this gives rise to a two-objective optimization problem. It is trivial to verify that CRPS and RS cannot simultaneously attain their minimum value. Indeed, by minimizing the former, $\eta_i = \frac{1}{2}\sqrt{\log 4}$ for any $i$ (see Eq. \ref{sigma_min}). Obviously, a constant $\eta_i$ cannot result in a minimum for RS, according to Eq. (\ref{optimal_eta}). Also, notice that trying to minimize RS as a function of $\sigma_i$ (for fixed errors $\varepsilon_i$) results in an ill-posed problem, because RS is solely expressed in terms of the relative errors $\eta$. Hence, there is no unique solution for the variances that minimizes RS. Hence, RS can be more appropriately thought of as a regularization term in the Accuracy-Reliability cost function. The simplest strategy to deal with multi-objective optimization problems is to scalarize the cost function, which we define here as
\begin{equation}\label{AR}
 \text{AR} = \beta\cdot \overline{\text{CRPS}} + (1-\beta)\text{RS}.
\end{equation}
We choose the scaling factor $\beta$ as
\begin{equation}\label{beta}
 \beta={\text{RS}}_{min}/(\overline{\text{CRPS}}_{min} + \text{RS}_{min}).
\end{equation}
The minimum of $\overline{\text{CRPS}}$ is $\overline{\text{CRPS}}_{min}=\frac{\sqrt{\log 4}}{2N}\sum_{i=1}^N \varepsilon_i$, which is simply the mean of the errors, rescaled by a constant. 
The minimum of RS follows from Eqs. (\ref{RS}) and (\ref{optimal_eta}): 
\begin{multline}
\text{RS}_{min} = \frac{1}{\sqrt{\pi} N}\sum_{i=1}^N \exp\left(-\left[\erf^{-1}\left(\frac{2i-1}{N}-1\right)\right]^2\right)\\
-\frac{1}{2}\sqrt{\frac{2}{\pi}}
\end{multline}
Notice that $\text{RS}_{min}$ is only a function of the size of the sample $N$, and it converges to zero for $N\rightarrow \infty$. 
The heuristic choice in Eq. (\ref{beta}) is justified by the fact that the two scores might have different orders of magnitude, and therefore we rescale them in such a way that they are comparable in our cost function (\ref{AR}). We believe this to be a sensible choice, although there might be applications where one would like to weigh the two scores differently.  Also, in our practical implementation, we neglect the last constant term in the definition (\ref{RS}), so that for sufficiently large $N$ $\text{RS}_{min}\simeq \frac{1}{2}\sqrt{\frac{2}{\pi}}\simeq 0.4$

\section{Results}
In summary, we want to estimate the input-dependent values of the empirical variances $\sigma_i^2$ associated to a sample of $N$ observations for which we know the errors $\varepsilon_i$. We do so by solving a multidimensional optimization problem in which the set of estimated $\sigma_i$ minimizes the AR cost function defined in Eq. (\ref{AR}). This newly introduced cost function has a straightforward interpretation as the trade-off between accuracy and reliability, which are two essential but conflicting properties of probabilistic models. In practice, because we want to generate a model that is able to predict $\sigma^2$ as a function of the inputs $\mathbf{x}$ on any point of a domain, we introduce a structure that enforces smoothness of the unknown variance, either in the form of a regularized Neural Network, or through a low-order polynomial basis. 
In the following we show some experiments with synthetic data to demonstrate the easiness, robustness and accuracy of the method.

\subsection{Experiments}
\begin{figure}[ht!]
\begin{center}
\centerline{\includegraphics[width=\columnwidth]{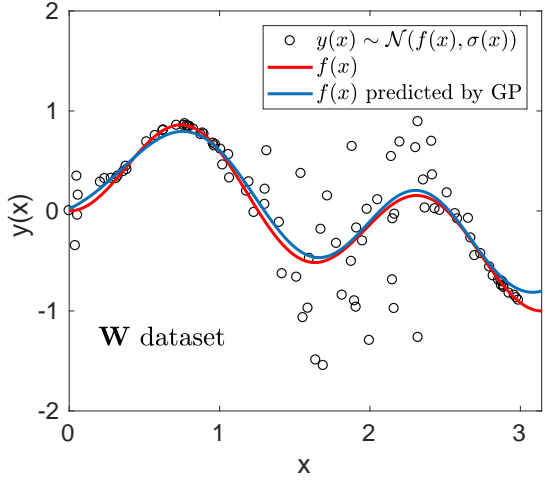}}
\caption{Same as in Figure \ref{G_regression}, for the {\bf W} dataset.}
\label{W_regression}
\end{center}
\vskip -0.1in
\end{figure}

\begin{figure}[ht]
\begin{center}
\centerline{\includegraphics[width=\columnwidth]{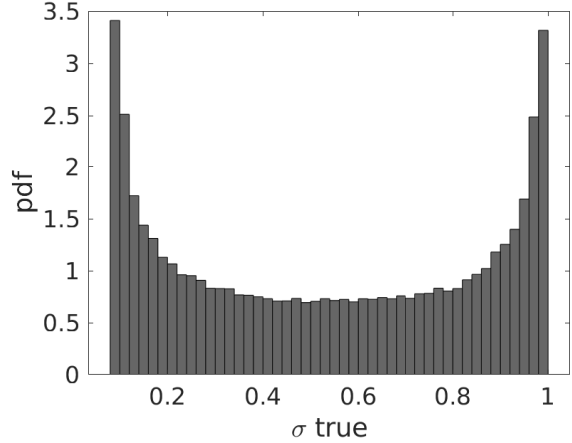}}
\caption{Distribution of true values of standard deviation $\sigma$ for the {\bf 5D} dataset.}
\label{multiD_2}
\end{center}
\vskip -0.1in
\end{figure}

In order to facilitate comparison with previous works, we choose some of the datasets used in \citet{kersting07}, although for simplicity of implementation we rescale the standard deviation so to be always smaller or equal to 1.
For all datasets the targets $y_i$ are sampled from a Gaussian distribution $\mathcal{N}(f(x),\sigma(x)^2)$. The first three datasets are one-dimensional in $x$, while in the fourth we will test the method on a five-dimensional space, thus showing the robustness of the proposed strategy.\\
{\bf G} dataset: $x \in [0,1]$, $f(x) = 2\sin(2\pi x)$, $\sigma(x) = \frac{1}{2}x+\frac{1}{2}$ \cite{goldberg98}. An example of 100 points sampled from the {\bf G} dataset is shown in Figure \ref{G_regression} (circles), along with the true mean function $f(x)$ (red), and the one predicted by a standard Gaussian Process regression model (blue).\\
{\bf Y} dataset: $x \in [0,1]$, $f(x) = 2(\exp(-30(x-0.25)^2)+\sin(\pi x^2))-2$, $\sigma(x) = \exp(\sin(2\pi x))/3$ \cite{yuan04}. An example of 100 points sampled from the {\bf Y} dataset is shown in Figure \ref{Y_regression} (circles), along with the true mean function $f(x)$ (red), and the one predicted by a standard Gaussian Process regression model (blue).\\
{\bf W} dataset: $x \in [0,\pi]$, $f(x) = \sin(2.5x)\sin(1.5x)$, $\sigma(x) = 0.01+0.25(1-\sin(2.5x))^2$ \cite{weigend94, williams96}. An example of 100 points sampled from the {\bf W} dataset is shown in Figure \ref{W_regression} (circles), along with the true mean function $f(x)$ (red), and the one predicted by a standard Gaussian Process regression model (blue).\\
{\bf 5D} dataset: $\mathbf{x} \in [0, 1]^5$, $f(\mathbf{x})=0$, $\sigma(\mathbf{x})=0.45(\cos(\pi + \sum_{i=1}^5 5x_i) + 1.2)$ \cite{genz84}.
Figure \ref{multiD_2} shows the distribution of $\sigma$, which ranges in the interval $[0.09,0.99]$.\\
For the {\bf G}, {\bf Y}, and {\bf W} datasets the model is trained on 100 points uniformly sampled in the domain. The {\bf 5D} dataset is obviously more challenging, hence we use 10,000 points to train the model (note that this results in less points per dimension, compared to the one-dimensional tests).
For all experiments we test 100 independent runs.

\subsection{Gaussian Process regression}
For the purpose of this work the specific method used to perform the regression problem on the mean function $f(\mathbf{x})$ is not crucial. However, for completeness we describe it here. We have used a standard Gaussian Process regression (that is, with constant homoskedastic noise), with squared exponential covariance:
\begin{equation}
 k(x_i,x_j) = \sigma_f^2\exp\left(-\frac{(x_i-x_j)^2}{2l^2}\right).
\end{equation}
The hyperparameters $\sigma_f$ and $l$ are found by minimizing the negative log likelihood on the training set.

\subsection{Neural Network}
For simplicity, we choose a single neural network architecture, that we use for all the tests. We use a network with 2 hidden layers, respectively with 20 and 5 neurons. The activation functions are $\tanh$ and a symmetric saturating linear function, respectively. The third (output) layer uses a squared exponential activation function, so the output is between 0 and 1. The dataset is randomly divided into training ($70\%$) and validation ($30\%$) sets. The network is trained using a standard BFGQ quasi-Newton algorithm, and the iterations are forcefully stopped when the loss function does not decrease for 10 successive iterations on the validation set. The only inputs needed are the points $x_i$ and the corresponding errors $\varepsilon_i$. The neural network outputs the values of $\sigma_i$, by minimizing the above-introduced Accuracy-Reliability cost function, Eq. (\ref{AR}). In order to limit the expressive power and avoid over-fitting, we add a regularization term equal to the L$_2$ norm of the weights to the AR cost function. This regularization term contributes 20\% to the total cost function. Finally, in order to avoid local minima due to the random initialization of the neural network weights, we train five independent networks and choose the one that yields the smallest cost function.

\subsection{Polynomial best fit}
In the case of low-dimensional data one might want to try simpler and faster approaches than a neural network, especially if smoothness of the underlying function $\sigma(x)$ can be assumed. For the one-dimensional test cases ({\bf G, Y, W}) we have devised a simple polynomial best fit strategy. 
We assume that $\sigma(x)$ can be approximated by a polynomial of unknown order, equal or smaller than 10: $\sigma(x)=\sum_{l=0}^{10} \theta_l x^l$, where in principle one or more $\theta_l$ can be equal to zero. The vector $\Theta=\{\theta_0,\theta_1,\ldots,\theta_{10}\}$ is initialized with $\theta_0=const$ and all the others equal to zero. The constant can be chosen, for instance, as the standard deviation of the errors $\varepsilon$. The polynomial best fit is found by means of an iterative procedure (Algorithm 1).

\begin{algorithm}[ht]
   \caption{Polynomial best fit}
   \label{alg:poly}
\begin{algorithmic}
   \STATE {\bfseries Input:} data $x_i, \varepsilon_i$
   \STATE Initialize $p = 0$, $\theta_0=const$, $P_{max}=10$, $tol$
   \WHILE{$p\leq P_{max}$ \& $err>tol$}
     \STATE ${p=p+1}$
     \STATE Initial guess for optimization $\Theta=\{\theta_0,\ldots,\theta_{p-1},0\}$
     \STATE $\Theta=\text{argmin AR}(\sigma_i)$ (with $\sigma_i = \sum_{l=0}^p \theta_lx_i^l$) 
     \STATE err = $||\text{AR}(\sigma(\Theta_{old})) - \text{AR}(\sigma(\Theta_{new}))||_2$
   \ENDWHILE 
\end{algorithmic}
\end{algorithm}
In words, the algorithm finds the values of $\Theta$ for a given polynomial order that minimizes the Accuracy-Reliability cost function. Then it tests the next higher order, by using the previous solution as initial guess. Whenever the difference between the solutions obtained with two successive orders is below a certain tolerance, the algorithm stops. The multidimensional optimization problem is solved by a BFGS Quasi-Newton method with a cubic line search procedure. Note that whenever a given solution is found to yield a local minimum for the next polynomial order, the iterations are terminated.

\begin{figure}[h]
\vskip 0.1in
\begin{center}
\centerline{\includegraphics[width=.8\columnwidth]{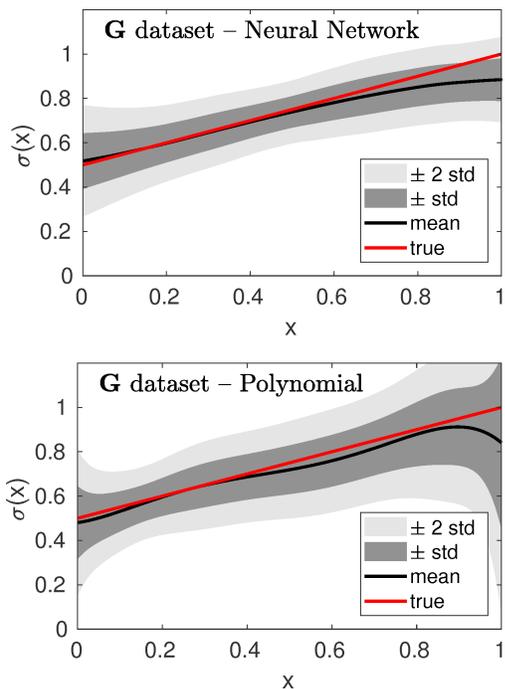}}
\caption{{\bf G} dataset: True value of the standard deviation $\sigma$ (red line) and mean value obtained averaging over 100 independent runs (black line). The gray shaded areas denote the confidence interval of one and two standard deviations calculated from the same ensemble of runs. In the top panel $\sigma$ is calculated through a Neural Network, while in the bottom panel as a polynomial function (see text). }
\label{G_dataset}
\end{center}
\vskip -0.1in
\end{figure}

\subsection{One-dimensional tests}
\begin{figure}[ht]
\vskip 0.1in
\begin{center}
\centerline{\includegraphics[width=.8\columnwidth]{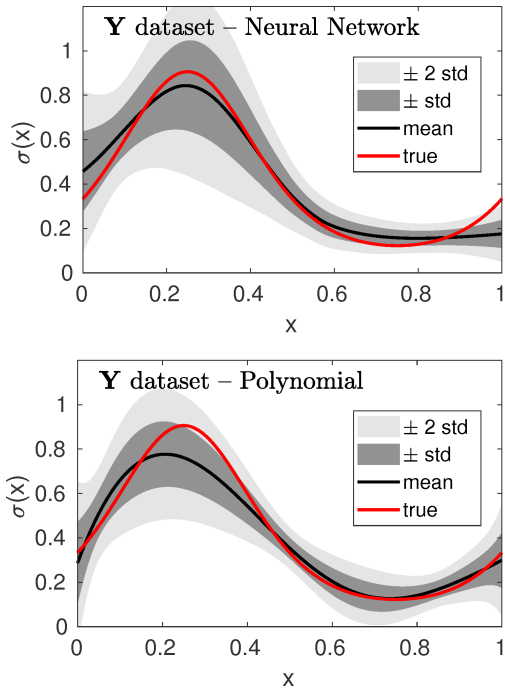}}
\caption{{\bf Y} dataset: True value of the standard deviation $\sigma$ (red line) and mean value obtained averaging over 100 independent runs (black line). The gray shaded areas denote the confidence interval of one and two standard deviations calculated from the same ensemble of runs. In the top panel $\sigma$ is calculated through a Neural Network, while in the bottom panel as a polynomial function (see text). }
x\label{Y_dataset}
\end{center}
\vskip -0.1in
\end{figure}

\begin{figure}[!ht]
\vskip 0.1in
\begin{center}
\centerline{\includegraphics[width=.8\columnwidth]{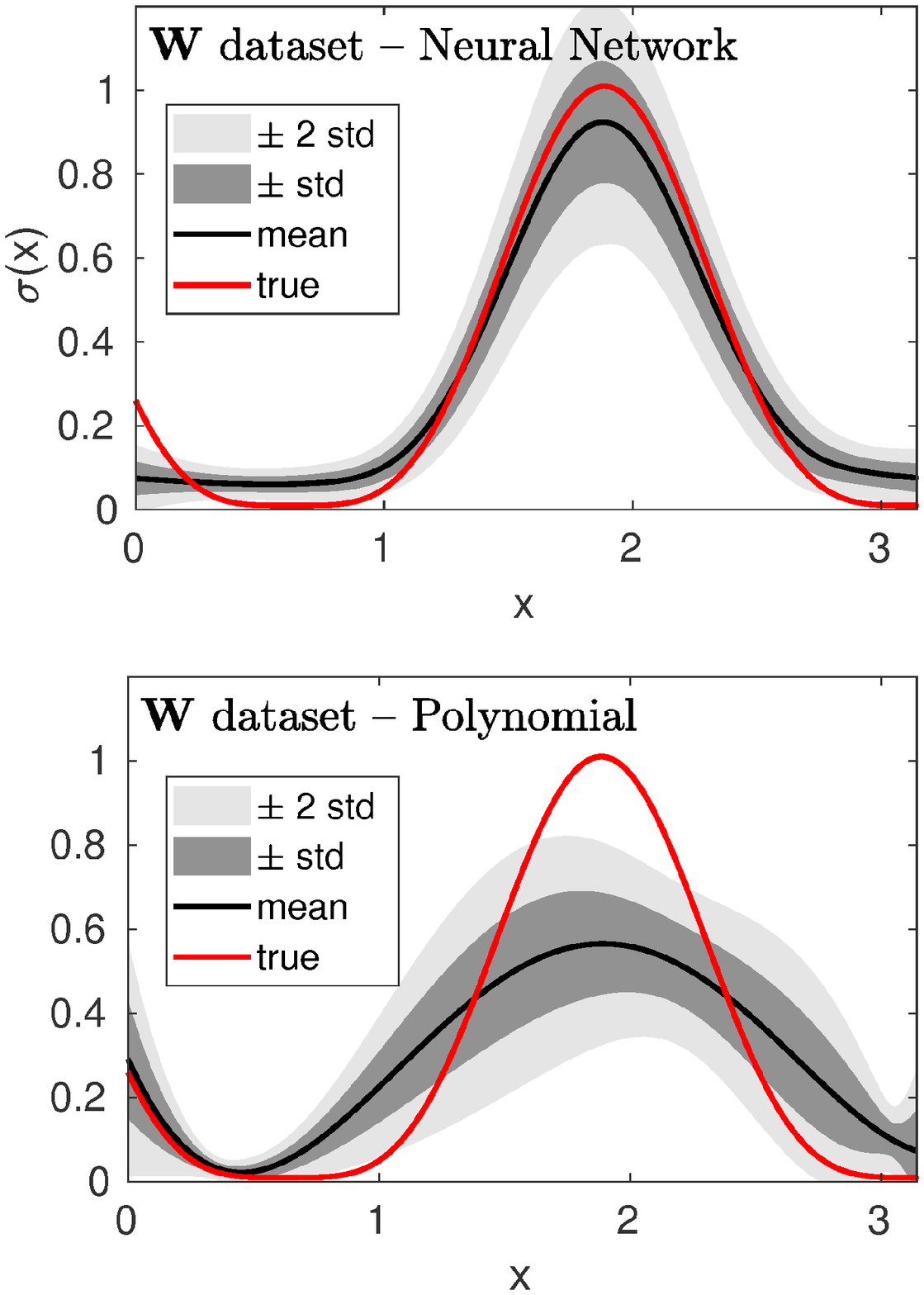}}
\caption{{\bf W} dataset: True value of the standard deviation $\sigma$ (red line) and mean value obtained averaging over 100 independent runs (black line). The gray shaded areas denote the confidence interval of one and two standard deviations calculated from the same ensemble of runs. In the top panel $\sigma$ is calculated through a Neural Network, while in the bottom panel as a polynomial function (see text). }
\label{W_dataset}
\end{center}
\vskip -0.1in
\end{figure}

The results for the 1D datasets $\bf G, Y, W$ are shown in Figures \ref{G_dataset} - \ref{W_dataset}, in a way consistent with \citet{kersting07}. The red lines denote the true standard deviation $\sigma(x)$ used to generate the data. The black line indicates the values of the estimated $\sigma$ averaged over 100 independent runs, and the gray areas represent one and two standard deviations from the mean. A certain spread in the results is due to different training sets (in each run 100 points are sampled independently) and, for the Neural Network, to random initialization.
The top panels show the results obtained with the Neural Network, while the bottom panels show the result obtained with the polynomial fit. In all cases, except for the {\bf W} dataset (polynomial case, bottom panel), the results are very accurate. 
We report in Table \ref{table:NLPD} the Negative Log Estimated Predictive Density $\text{NLPD} = -1/n\sum \log p(y_i)$ calculated over $n=900$ test points, where $p_i$ is the probability density of a normal distribution with mean $\mu_i$ and standard deviation $\sigma_i$. 
We have calculated the NLPD for 100 independent runs and we report in the Table the values of the quartiles (due to the presence of large outliers, these are more representative than mean and standard deviation). We compare the results obtained with a standard (homoskedastic) GP, our Neural Network implementation, and the polynomial fit implementation, clearly showing the strength of the new method.

\subsection{Multi-dimensional test}
\begin{figure}[t]
\begin{center}
{\includegraphics[width=.8\columnwidth]{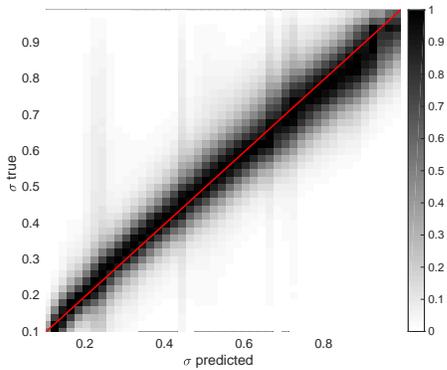}}
\caption{Probability density of the prediction versus real values of $\sigma$ for the {\bf 5D} dataset. The red line denotes perfect prediction. The densities are normalized to have maximum value along each column equal to one. 10,000,000 samples have been used to generate the plot (with a training set of 10,000 points).}
\label{multiD_1}
\end{center}
\end{figure}

For the {\bf 5D} dataset it is impractical to compare graphically the real and estimated $\sigma(\mathbf{x})$ in the 5-dimensional domain. Instead, in Figure \ref{multiD_1} we show the probability density of the real versus predicted values of the standard deviation. Values are normalized such that the maximum value in the colormap for any value of predicted $\sigma$ is equal to one (i.e. along vertical lines). The red line shows a perfect prediction. The colormap has been generated by 10e6 points, while the model has been trained with 10,000 points only.
For this case, we have used an exact mean function (equal to zero), in order to focus exclusively on the estimation of the variance.
We believe that this is an excellent result for a very challenging task, given the sparsity of the training set, that shows the robustness of the method.

\begin{table}[t]
\caption{Negative Log Estimated Predictive Density}
\label{table:NLPD}
\vskip 0.15in
\begin{center}
\begin{small}
\begin{sc}
\begin{tabular}{lcccc}
\toprule
Dataset &  &GP &AR-NN& AR-Poly \\
\midrule
      &1$^\text{st}$ quartile      & 1.22 & 1.21 & 1.15 \\
{\bf G} & median                   & 1.28 & 1.26 & 1.17 \\
      &3$^\text{rd}$ quartile      & 1.37 & 1.33 & 1.21 \\
\midrule
      &1$^\text{st}$ quartile      & 1.23 & 0.49 & 0.40 \\
{\bf Y} & median                   & 1.56 & 0.57 & 0.44 \\
      &3$^\text{rd}$ quartile      & 2.16 & 0.65 & 0.51  \\
\midrule
      &1$^\text{st}$ quartile      & 1.15 & -0.15 & 0.06\\
{\bf W} & median                   & 1.68 & -0.03 & 0.17\\
      &3$^\text{rd}$ quartile      & 3.18 & 0.09 & 0.37 \\      
\bottomrule
\end{tabular}
\end{sc}
\end{small}
\end{center}
\vskip -0.1in
\end{table}

\section{Discussion and future work}
We have presented a simple parametric model for estimating the variance of probabilistic forecasts or, equivalently, to compute the Gaussian noise in heteroskedastic regression.
We assume that the data is distributed as $\mathcal{N}(f(\mathbf{x}),\sigma(\mathbf{x})^2)$, and that an approximation of the mean function $f(\mathbf{x})$ is available (the details of the model that approximates the mean function are not important). In order to generate the uncertainty $\sigma(\mathbf{x})$, we propose to minimize the Accuracy-Reliability cost function, which depends only on $\sigma$, on the errors $\varepsilon$, and on the size of the training set $N$. We have discussed how accuracy and reliability are two conflicting metrics for a probabilistic forecast and how the latter can serve as a regularization term for the former. We have shown that by solving this multidimensional optimization problem, one is able to discover, with high accuracy, the hidden noise function. An important point to notice is that by setting the problem as an optimization for the Accuracy-Reliability cost function, the result for $\sigma$ will be optimal, for the given approximation of the mean function $f(\mathbf{x})$. In other words, the method will inherently attempt to correct any inaccuracy in $f(\mathbf{x})$ by assigning larger variances.  Hence, the agreement between predicted and true values of the standard deviation $\sigma$ presented in Figures \ref{G_dataset}-\ref{W_dataset} must be understood within the limits of the approximation of the mean function provided by the Gaussian Process regression.
By decoupling the prediction of the mean function (which can be assumed as given by a black-box) from the estimation of the variance, this method results in a much lower cost than strategies based on a Bayesian framework (and subsequent approximations). Moreover, for the same reason this method is very appealing in all applications where the mean function is necessarily computed via an expensive black-box, such as computer simulations, for which the de-facto standard of uncertainty quantification is based on running a large (time-consuming and expensive) ensemble.
Finally, the formulation is well suited for high-dimensional problems, since the cost function is calculated point-wise for any instance of prediction and observation.\\
Although very simple and highly efficient the method is fully parametric, and hence it bears the usual drawback of possibly dealing with a large number of choices for the model selection. An interesting future direction will be to incorporate the Accuracy-Reliability cost function in a non-parametric Bayesian method for heteroskedastic regression.

\section*{Acknowledgements}
All the codes used to generate the data will be made available on our website \texttt{www.anonymized.com}



\end{document}